# A Neuro-Symbolic Humanlike Arm Controller for Sophia the Robot

David Hanson[1], Alishba Imran[2], Abhinandan Vellanki[3], Sanjeew Kanagaraj[4]

Hanson Robotics Ltd

david@hansonrobotics.com[1], alishba.imran.ext@hansonrobotics.com[2], abhinandan.vellanki@hansonrobotics.com[3], sanjeew.kanagaraj@hansonrobotics.com[4]

**Abstract**

We outline the design and construction of novel robotic arms using machine perception, convolutional neural networks, and symbolic AI for logical control and affordance indexing. We describe our robotic arms built with a humanlike mechanical configuration and aesthetic, with 28 degrees of freedom, touch sensors, and series elastic actuators. The arms were modeled in Roodle and Gazebo with URDF models, as well as Unity, and implemented motion control solutions for solving live games of Baccarat (the casino card game), rock paper scissors, hand shaking, and drawing. This included live interactions with people, incorporating both social control of the hands and facial gestures, and physical inverse kinematics (IK) solving for grasping and manipulation tasks. The resulting framework is described as an integral part of the Sophia 2020 alpha platform, which is being used with ongoing research in the authors' work with team AHAM, an ANA Avatar Xprize effort towards human-AI hybrid telepresence. The uses of the work extend across domains, and include arts and social human-robot interaction, as well as targeting more general co-bot applications. These results are available to test on the broadly released Hanson Robotics Sophia 2020 robot platform, for users to try and extend.

## Grasping the Situation

Although recent AI and intelligent robotics are increasingly effective in many applications, they are often narrowly focused and effective only on domain-specific use cases. For example, while robotic grasping and manipulation have matured in both research and deployed applications (Liu, 2020), the results are narrow, the associated AI is not as adaptive or understanding as those of humans. The pursuit of human-level general intelligence remains a distant goal, as today's machines are still brittle and narrow in capabilities (Goertzel, 2014).

In the pursuit of new kinds of AI, the biology of human cognition may serve as a useful reference (Adolphs, 2001), and to this end, here we describe a platform and research that strives towards a useful approximation of some elements by which humans cognition arises from physical embodiment within a physical environment, by combining humanlike robotic grasping and social abilities, with symbolic AI with explicit knowledge models and inference, and deep learning networks, in an adaptive, reconfigurable framework.

With the described work, we strive to a particular approximation of features, including embodiment with motorized actuation and sensor data and various modes of perception, social agency for human interactions, cognitive software features including working memory, long term and short term memory, inferential reasoning, and some software approximation evolutionary psychology providing drives, urges and motives. While these features are included in a preliminary form, we think that they can be used in learning and goal pursuit, and enough platform flexibility to allow researchers to explore different technology configurations, for many future experiments.

## Background

The described work project is the latest step in developing a conversationally interactive character called Sophia 2020. Sophia is a robot designed by Hanson Robotics to serve as a humanlike agent with robotics embodiment to produce the minimal channels of feedback for an agent to able to learn it's environment, both with symbolic AI and neural networks. She is also part of Team AHAM, hybrid AI-human controlled telepresence robot, a collaboration of IISc, TCS, Tata, and Hanson Robotics on the ANA Avatar X-Prize. With Sophia, we combine a wide variety of AI and robotic technologies to create physically embodied science and fiction, as a kind self-referential meta-fiction, bringing practices from AI agents with interactive game character design, together with robotics, as a new conceptual art. Building on the legacy of previous work with android portraits of Philip K Dick, Bina Aspen Rothblatt (Bina 48), Zeno, and others, which depicted living people with autonomous, intelligent androids (Hanson, 2014).

In a typical interaction scenario, Sophia interacts with people socially. Face detection software (PoseNet, and OpenCV) will detect a person, and the robot will be controlled to make eye contact and smile in greeting. Automatic speech recognition (Google and Open Speech) will detect users' speech input, transcribed as text, and send this as text to the natural language processing core.

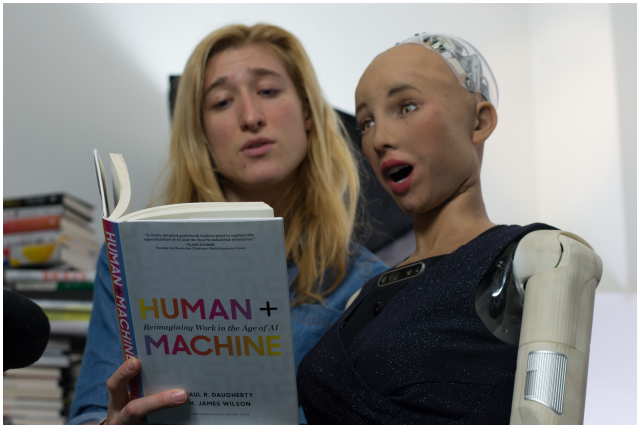

Figure 1, Sophia the Robot and a human friend

The determined response will then drive the facial animation, controlled by integration with Blender and ROS, with in sync with a highly realistic synthesized TTS voice custom trained, provided by Cereproc. The realistic facial expressions are generated using Hanson Robotics lifelike skin material Frubber to affect naturalistic expressions that simulate the major 48 muscles of the human face, including all the facial action units in the facial action coding system (FACS), with lower power and greater verisimilitude than other materials (Hanson, 2005, Hanson 2009; Bar-Cohen and Breazeal, 2003). The lightweight low-power characteristics of the hardware make it appropriate for untethered bipeds, as demonstrated on the walking Sophia-Hubo (Park, 2018), and is appropriate for mass manufacturing.

**An Integrative Platform for Embodied Cognition.**

We worked to build the experiments in cognitive arm and grasping controls upon the latest experimental version of the Hanson Robotics framework for human-like embodied cognition, called Sophia 2020. This platform brings together humanoid robotics hardware with expressive humanlike face, gestural arms, locomotion, and a is designed as a toolkit integrating a neuro-symbolic AI dialog ensemble, novel robotic hardware with humanlike expressive face for social learning and communications, robotic arms and locomotion, and a wide array of sensors, machine perception, and motion control tools. Here we present both the technology architecture as an alpha platform release, and experimental results in a variety of tests including human-robot interaction, ensemble verbal and nonverbal dialogue interactions, and mechanical tasks such as facial and arms controls, and applications in the arts, therapeutic healthcare, and telepresence. While many cognitive robotics frameworks exist, the Sophia 2020 framework provides unique expressive animations in facial gestures, character agent authoring tools, open interfaces, and a diversity of features and extensions.

The framework builds on prior work building previous robots and cognitive systems, including: Bina-48, the PKD android, the INDIGO cognitive robots Alice and Aleph (Zerrin, 2012), Zeno (Hanson, 2010), and others. Combining results from these previous works within a new mass-manufacturable hardware and software framework, Sophia 2020 is designed as a platform for cognitive robotics, with an ensemble of the minimum components that are speculated to be related to complex, factors to result in human-level cognition, under the conjecture that consciousness is a multidimensional phenomenon requiring physically embodiment.

**Arm and Hand Controls**

The foundation of the arm and hand controls combined classic motion control with computer animation, and an open-ended direct motor control interface, wrapped in an SDK and API integrated with the Robot Operating System (ROS). The arms and hands' mechanical models and mechanical properties were used to build URDF models, used in several motion control frameworks (Roodle, Gazebo, and prototyped in Moveit), demonstrating both position feedback and force feedback controls, IK solvers, and PID loops. We also considered how more bio-inspired motor controls could be created with the compliant series elastic actuators and programmatic compliance, with central pattern generators; however these specs are in a design phase, and remain to be integrated with existing models of such controls approached.

**Motion Controls**

To facilitate appealing, naturalistic motion control, we turned to animation controls, as used in procedural and programmatic animations typical in videogames. For the infrastructure, we created a shared control schema that integrates open source Blender 2.0 and allows for control from Blender rig, to be blended with the robotic motor control framework, in shared control, and then subsumed by the utility based motor control. This allows for the character animations, a.k.a. "illusion of life", and practical interaction with the physical world.

For the making of the character actions, the overall performance was made by combining Face and Arm animations that were based motion capture data and handcrafted animation. A Virtual Sophia rig was exported into unity, and used to provide realtime character simulation used to visualize and author the test animations.

For motion capture data, we developed procedural retargeting tool to map animations to current model of robot. For facial motion capture data, we used Apple ARkit Technology on IOS devices. Developed capabilities to live stream and retarget facial motion capture data directly to the robot face. To achieve lip sync performance, we used TTS engine provided visual phonemes (visemes) with timings for each viseme.

For automated, generative animated character actions, we started with a library of facial and arm gestures. That library we used based on the preselected keywords and dialog act for given sentences. So the robot can made corresponding gestures based on the content of the sentence, and then added stochastic noise, variable amplitudes, and motion blending, as well as wave generations to simulate the signals of the CPG in the human motion generation.

For the hardware design and first motion control approach, we first built arms in 2014, with a version of arms developed for gestures and card game baccarat dealing. The unique challenges for this application:

**Baccarat Dealer Application**
The early special version of the arms were developed for card game baccarat dealing. Those
The unique challenges for this application:

- Robot should deal 4 base cards and additional 2 cards (if necessary) in less than half a minute. In a two games per minute pace.
- Pass cards between hands for better efficiency. Demo: https://drive.google.com/file/d/1Hrukt5lN5wTo_augULaMZ4LIDWVA8j7s/edit

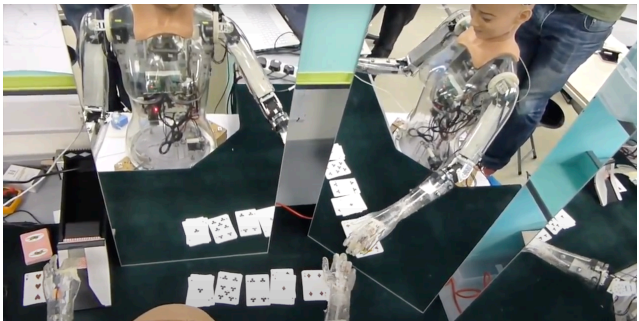

Figure 2, Sophia dealing out cards.

We approached the problem by utilizing constraints based multi-task inverse kinematic engine for the whole robot body - "Manypulator". The whole robot was described as a kinematic chain and multiple tasks allowed to execute for different end end-effectors with the constraints applied to specific joints and the synchronization being required between multiple tasks. While at first we haven't met our performance requirements, we had developed visual debugging and authoring tools that allowed us to fine tune all motion parameters and achieve an average drawing speed of 25s for 6 cards dealt. We also achieved 99.5% drawing and placing accuracy with the tuned speed with certified casino equipment.

**Roodle - Ros Doodler**
The other application we developed for later iteration of the arms were drawing human faces and other objects acquired by cameras, or generated using GAN neural networks trained from the images acquired from robot cameras. The pipeline of this task had following steps:
 a. Acquire image
 b. Find the foreground object and remove all other background.
 c. Contrast and brightness adjustment based on the preferred level of details of the drawing was required.
 d. Using algorithm developed in collaboration with artist Patrick Tresset, convert the lines and curves to the image

The curves then would be segmented into short lines and merged with lines and through ROS interface sent to the inverse kinematics engine.

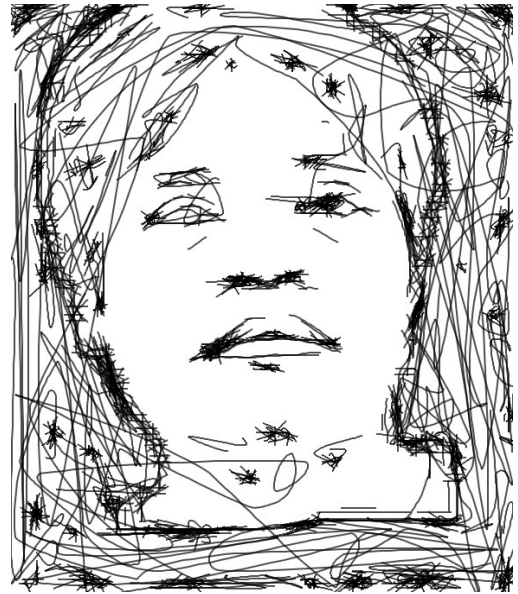

Figure 3, Sophia drawing generated and drawn with robotic arm.

The approach we then took was developing a ROS bridge on top of the Manypulator inverse kinematics engine. In addition tools for calibration simulation and tuning of line drawing were implemented and the whole application was named Roodle. While to define the plane in 3D space only 3 point calibration is necessary, we used 5 points to calibrate the plane to account for small inaccuracies and play in the actuators.

With calibrated pain, we detached the Wrist pitch (down) actuator from the kinematic chain in order to use it for the pen handling. We would limit the torque of this actuator on its way down, to avoid pen hitting paper, or overshooting starting point. The torque of the wrist would be increased, during rest of the motion, to make the lines straight.

While our approach was designed to work in the various settings with only approximate calibration possible at times, the redundancies we implemented help to have accurate drawings even with small changes in drawing plane position without need of recalibration.

For future work we consider hand eye-hand coordination for automatic calibration and parameters optimization, together with grasping and manipulating pen.

The simple architecture of approach:

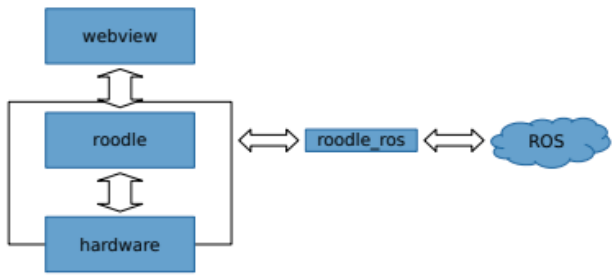

Roodle architecture::

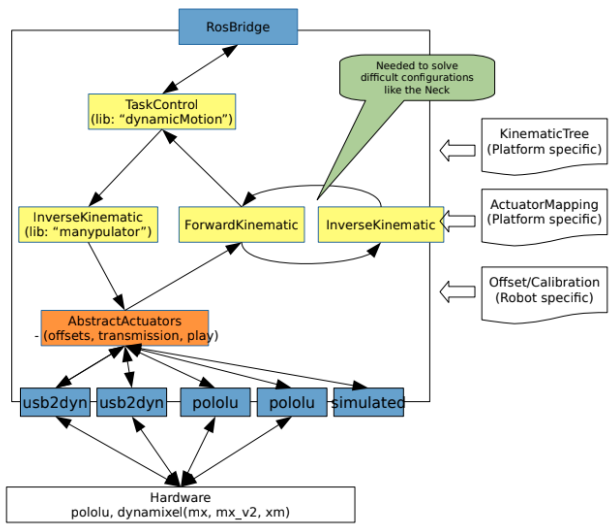

Figure 4, Sophia's arm control architecture.

Mechanically, we developed our own robotic arms, to suit the character and collaborative robotic needs. We used the dimensions from a medium-sized woman's arms as constraints and quality components integrated within the robots. We try to have feedback in all the DOF axes where possible.

The principal structure of the arms are made of CNC Aluminum 6061, and for complicated manufactured parts are made of Nylon SLS PA 2200, black color for aesthetic. Each arm has 14 DoF (7 for the arm and wrist, and 7 in the hands) and uses Dynamixel servos for controlling and for feedback.

Because of the size limitation, each hand uses XPERT CI-2401 and MKS_HV75K with Maestro Controller 24ch. For feedback, there are pressure sensor pads and potentiometers in each finger and both wrists.

The goal of Sophia arms is for gesticulation when she speaks and also for object grasping & manipulation.

**Arm Axes**

The arms use Dynamixel servos and the selection of those servos are determined by the weight, torque and space limitations. The torque of each Axes was calculated in Solidworks with the center of mass and the weight ( Table 1)

The arms use Dynamixel servos and the selection of those servos are determined by the weight, torque and space limitations. The torque of each Axes was calculated in Solidworks with the center of mass and the weight ( Table 1).

The features of Dynamixel servos are: Smart actuator with fully integrated DC Motor + Reduction Gearhead + Encoder + Controller + Driver. Functions include precise control, PID control, 360 degree of position control, and high speed communication. For the Wrist Roll and Pitch there are torsion springs for safety use.

**Hand Axes**

Because of the space limitations, there are 3 Xperts servos and 1 MKS servo in the hand. There are 3 Xperts in the forearm, which control the fingers and the spreading. (Table 2). For the feedback, each fingertip has an internal pressure sensor pad with capacity of 500gr; the sensor is activated through a Nylon-rubber fingertip cover.

There are six 10K ohms potentiometers for measuring the position of 4 fingers and 2 for the wrist. For index, middle ring, pinky fingers and thumb, there are torsion springs in one direction for safety use.

| Axes | Weight to handle (gr) | Driven by | Ratio | Max Pulley Torque | Max Motor Torque | Motor | Feedback |
|---|---|---|---|---|---|---|---|
| Shoulder Pitch | 2605 | Pulley | 2:1 | 6.451 Nm @15rpm(90deg/seg) | 3.226 Nm @30rpm(180deg/seg) | MX106R | Position, speed, Acceleration, temperature in motor |
| Shoulder Roll | 2322 | Pulley | 2:1 | 6.476 Nm @15rpm(90deg/seg) | 3.238 Nm @30rpm(180deg/seg) | MX106R | Position, speed, Acceleration, temperature in motor |
| Shoulder Yaw | 1830 | Spur Gears | 2.917 : 1 | 1.967Nm@13.16rpm(79deg/seg) | 0.674Nm@38.4rpm(230.4deg/seg) | XM430-W350 | Position, speed, Acceleration, temperature in motor |
| Elbow | 1278 | Pulley | 2:1 | 2.018 Nm @26.5rpm(159deg/seg) | 1.009 Nm @53rpm(318deg/seg) | MX64R | Position, speed, Acceleration, temperature in motor |
| Wrist Yaw | 946 | Bevel Gears | 1.8 : 1 | 0.315 Nm @25rpm(150deg/seg) | 0.175Nm @45rpm(270deg/seg) | XM430-W350 | Position, speed, Acceleration, temperature in motor |
| Wrist Roll | 384 | Pulley | 1.33 : 1 | 0.625Nm@30.8rpm(185deg/seg) | 0.469Nm@41rpm(246.6deg/seg) | XM430-W350 | Position, speed, Acceleration, temperature in motor+ Potentiometer in axis |
| Wrist Pitch | 371 | Pulley | 1.467 : 1 | 0.501Nm@29.18rpm(175.1deg/seg) | 0.342Nm@42.8rpm(256.81deg/seg) | XM430-W350 | Position, speed, Acceleration, temperature in motor+ Potentiometer in axis |

| Hand Axis | Drive | Torque (Kgcm)@Speed (sec/60degre) | Motor | Feedback |
|---|---|---|---|---|
| Index | Series elastic | 6.02@0.069@7.4V | XPERT_CI-2401-HV | Spring potentiometer series elastic actuator (SEA) + Force FSR touch |
| Middle | Series elastic | 6.02@0.069@7.4V | XPERT_CI-2401-HV | Potentiometer (SEA) + FSR touch sensor |
| Ring | Series elastic | 6.02@0.069@7.4V | XPERT_CI-2401-HV | Potentiometer (SEA) + FSR touch sensor |
| Pinky | Series elastic | 6.02@0.069@7.4V | XPERT_CI-2401-HV | FSR touch sensor |
| Thumb | String | 6.02@0.069@7.4V | XPERT_CI-2401-HV | Potentiometer (SEA) + FSR |
| Thumb Roll | String | 6.02@0.069@7.4V | XPERT_CI-2401-HV | None |
| Spreading | Rigid linkage | 2.8@0.1@7.4V | MKS_HV75K | None |

Table 1: Sophia Arm mechanical properties

.

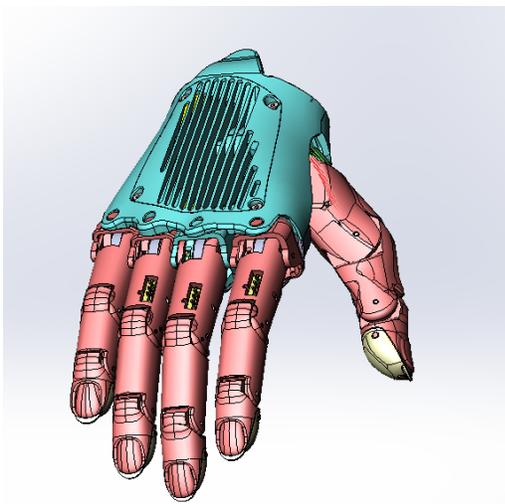

Figure 5, Sophia hand design.

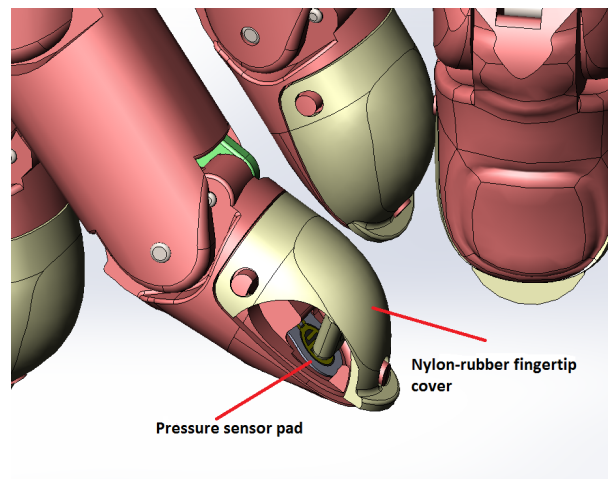

Figure 6, Force sensing fingertips in Sophia 2020 platform.

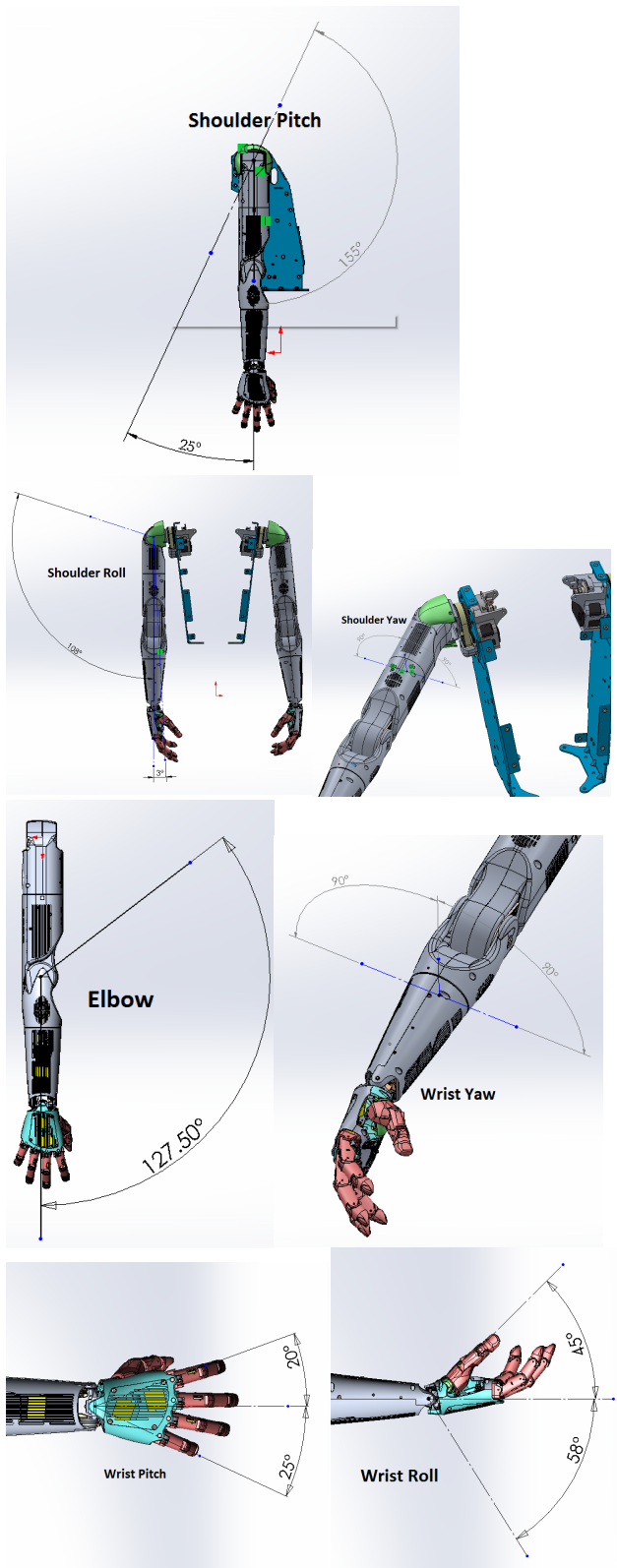

Figure 7, Range of motion Sophia arms and hands.

**Electrical**

The power of the Arm is 13.8V, stepped down from 24V, and the communication is with RS485 in chain for Dynamixel. The power of the hand is 7.4V, stepped down from 24V, and the communication is using PMW (pulse width modulation) through a 24 channel Maestro-micro-controller.

**Algorithms and Controls**

The algorithm used for grasping detection is a Generative Grasping Convolutional Neural Network (GG-CNN) widely inspired by the "Closing the Loop for Robotic Grasping: A Real-time, Generative Grasp Synthesis Approach" paper. We built on this algorithm by testing it on a robotic human-like hand and increasing the grasping tasks. A GG-CNN works by taking in depth images of objects and predicting the pose of grasps at every pixel for different grasping tasks and objects.

CNN-based controllers for grasping are preferred since they can do closed-loop grasping. Both systems learn controllers which map potential control commands to the expected quality of or distance to a grasp after execution of the control, requiring many potential commands to be sampled at each time step.

- Currently classifying grasp candidates are sampled from an image or point cloud, then are ranked individually using a CNN. Once the best grasp candidate is determined, a robot executes the grasp open-loop (without any feedback).
- This requires precise calibration between the camera and the robot, precise control of the robot and a completely static environment. This can also take a long time and lots of computation to run in real-time.
- **GG-CNN is better:** Using this we can directly generate grasp poses for every pixel in an image simultaneously, in a closed-loop manner which also uses a comparatively small neural network.

**What is GG-CNN?**
- Takes in real-time depth images and identifies objects through object detection.
- Parameterized as a grasp quality, angle and gripper width which is done for every pixel in the input image in a fraction of a second.
- The best grasp is calculated and a velocity command (v) is issued to the robot.

**Benefits:**
- Less compute since less parameters are used.

- Grasping task is done pixel by pixel which is faster and more accurate.
- Closed-loop (takes in feedback from the previous batch to improve overtime). Contains visual servoing which are able to adapt to dynamic environments and do not necessarily require fully accurate camera calibration or position control.

**Grasping Pose**

Grasp is executed perpendicular to a plane surface, given a depth image of the scene which is determined by its pose (such as the grippers Centre position based on x,y,z values in Cartesian coordinates). The grippers rotation around the z axis and the required gripper width. A scalar value is also used to represent the chances of grasp success in the pose.

A grasp can be described as:

$$\tilde{\mathbf{g}} = (\mathbf{s}, \tilde{\phi}, \tilde{w}, q)$$

- $\mathbf{s}$ = (u, v) which is the centre point in pixels.
- $\tilde{\phi}$ is the rotation in the camera's reference frame.
- $\tilde{w}$ is the grasp width in pixels.

A grasp in the image space can be converted to a grasp in world coordinates by transforming it with the following:

$$\mathbf{g} = t_{RC}(t_{CI}(\tilde{\mathbf{g}}))$$

- $t_{RC}$ transforms from the camera frame to the world/robot frame.
- $t_{CI}$ transforms from 2D image coordinates to the 3D camera frame, based on the camera's parameters and defined calibration between the robot and camera.

**Grasp Representation:** represents a grasp map (G) as a set of three images, Q, Φ and W:
- Q is an image that describes the quality of a grasp executed at each point (u, v). The value is a scalar in the range [0, 1] where a value closer to 1 indicates higher grasp quality.
- Φ is an image that describes the angle of a grasp to be executed at each point. The angles are given in the range [−π/2, π/2].
- W is an image that describes the gripper width of a grasp to be executed at each point. To allow for depth invariance, values are in the range of [0, 150] pixels, which can be converted to a physical measurement using the depth camera parameters and measured depth, and the gripper used.

Dataset: Cornell Grasping

To train our network, we used the Cornell Grasping Dataset which contains 885 RGB-D images of real objects, with 5110 human-labelled positive and 2909 negative grasps. This dataset is good for our task because it has a pixelwise grasp representation as multiple labelled grasps. We also captured

Depth and RGB images from the Cornell Grasping Dataset with the ground-truth positive grasp rectangles are shown in green. From the ground-truth grasps, the Grasp Quality (QT), Grasp Angle (ΦT) and Grasp Width (WT) images are generated to train the network. The angle is further decomposed into cos(2ΦT) and sin(2ΦT) for training.

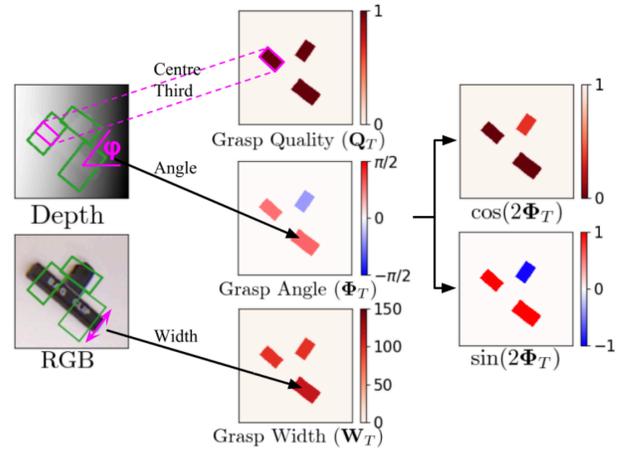

Figure 8, Grasp Poses Process Overview

**Results:**

Based on current tests, the model is able to get a 87% grasp success rate on a set of previously unseen objects and 92% on a set of household objects that are moved during the grasp attempt and 90% accuracy when grasping in dynamic clutter. This model achieved 5.3% and 20.4% higher grasp accuracy compared to the ENet [Paszke et al., 2016] and MobileNet [Howard et al., 2017] models, respectively.

**Reinforcement Learning & Imitation Learning Approaches**

Reinforcement Learning (RL) figures out what to do and how to map situations to actions. The end result is to maximize the numerical reward signal but instead of telling the learner what action to take, they must discover which action will result in the maximum reward. In our case, the action would be grasping an object with a high success rate.

Using RL, we can get an agent to learn the optimal policy for performing a sequential decision without complete knowledge of the environment. The agent first explores the environment by taking action and then edits the policy according to the reward function to maximize the reward. To train the agent, we can use:

- **Deep Deterministic Policy Gradient (DDPG).** A model-free off-policy algorithm for learning continous actions.
- **Trust Region Policy Optimization (TRPO).** Policy Gradient methods (PG) are commonly used and at a high level use gradient ascent to follow policies with the steepest increase in rewards. This is not very accurate for curved areas though and that is why TRPO are effective for optimizing large non-linear policies.
- **Proximal Policy Optimization (PPO).** At a very high level, PPO have some of the benefits of TRPO but are much simpler to implement and tune.

All algorithms are variations of the deep neural network architecture shown in Figure 7 to represent the Qfunction.

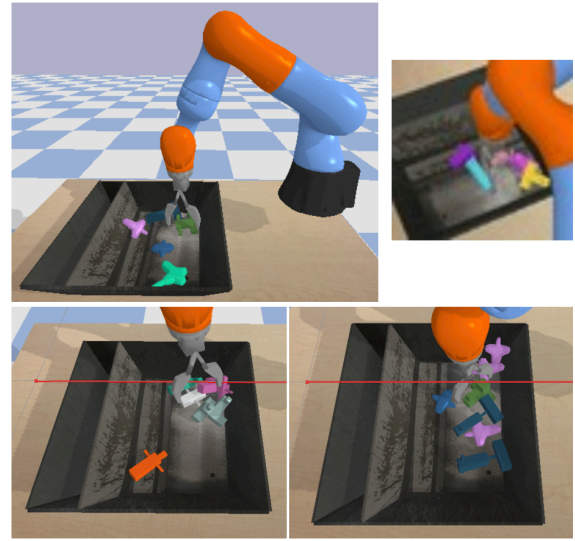

Figure 10, Simulation learning with the algorithm. The robot picks up a wide variety of randomized objects and generalizes to unseen test objects

**Neuro-Symbolic AI**

The long-term the goal is to integrate it with symbolic reasoning techniques such as rules engines or expert systems or knowledge graphs. This is often referred to as Nero-Symbolic AI. Examples of this can be using neural networks to identify what kind of shape or colour a particular object has and then applying symbolic reasoning to identify other properties such as the area of the object, volume and so on.

**Why should we combine neural networks with symbolic reasoning?**

Current deep learning models are too narrow. When you give them huge amounts of data, they work very well at the task you want it to perform but break down if you prompt it to adapt to a more general task. To do this you also need enormous amounts of data. On the other hand, symbolic AI is really good at doing interesting things with symbols but actually getting the symbols from the real world is much harder than anticipated.

By combining the two approaches, you end up with a system that has neural pattern recognition allowing it to see, while the symbolic part allows the system to logically reason about symbols, objects, and the relationships between them.

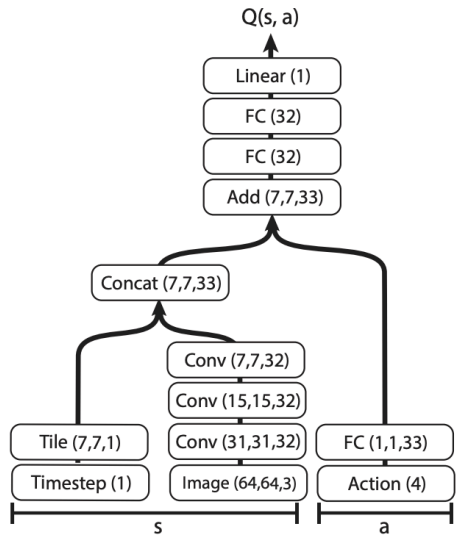

Figure 9, Q network architecture used for grasping tasks. The model takes as input the timestep t, the image observation s, and candidate action a.

In our set up, our robotic arm attempts to grasp objects from a bin. The arm has a fixed number of timesteps (T = 10) in which it has to find the best grasp at which point the gripper closes and the episode ends. At the beginning of each new episode, the object positions and rotations are randomized within the bin. The reward is binary and provided at the last step, 1 for a successful grasp and 0 for a failed grasp. The observed state consists of the current RGB image from the viewpoint of the robot's camera and the current timestep t which is included in the state, since the policy must know this before making a decision.

The robot must pick up objects in a bin, which is populated using randomized objects and then generalize to unseen test objects or pick up a specific object from a cluttered bin.

How Neuro-Symbolic AI works:
- A neural network for object detection can be used to map from inputs (like an image of an apple) to output (like the label "apple").
- Symbolic AI is different since it would instead express all the knowledge we have about apples: an

apple has parts (a stem and a body), it has properties like its color, it has an origin (it comes from an apple tree), and so on.
- Combining these two approaches will allow you to still detect the object but draw more insights about the object, the environment it's in and it's interactions.

MIT-IBM CLEVRER

CLEVR is one of the first datasets using neuro-symbolic ai that provides a set of images that contain multiple solid objects and poses questions about the content and the relationships of those objects (visual questioning). If you want to train a neural network to solve a system like that, you might be able to do it but it requires tremendous amounts of data.

To solve this, they first use a CNN to process the image and create a list of objects and their characteristics such as color, location and size. These are types of symbolic representation that rule-based AI models can be used on. Another NLP algorithm processes the question and parses it into a list of classical programming functions that can be run on the objects list. The result is a rule-based program that expresses and solves the problem in symbolic AI terms as shown in Figure 8. The AI model is trained with reinforcement learning through trial and error based on the rules of the environment it is operating in.

**How could we do this for grasping?**

We can create a map of different applications and capabilities through affordances. Affordances include the perceived and actual properties of a thing that provide cues to the operation of things. Affordances can be symbolic representations used to complete a task. Combining these capabilities with neural network which can make a prediction for grasping and turn that into symbolic representation that can be reasoned with through calling functions. Although we still have to further implement this, these are some of the ideas that our team is now working on testing.

**Next Steps**

Our team is working on the following to progress this project forward:
- Test out current gg-cnn model with ROS implementation on robotic arm.
- Generate our own dataset for applications that Sophia will be tested on.
- Create model for multi-finger grasp.
- Going deeper into Neural-Symbolic AI and framing how we can use this technique for manipulation tasks such as grasping.
- Integrate model into Hanson Robotics simulation environment to test out model on Sophia. Eventually, the goal is to use the model alongside existing systems that are based on symbolic representations.
- Port to ROS 2.0 and Moveit 2.0

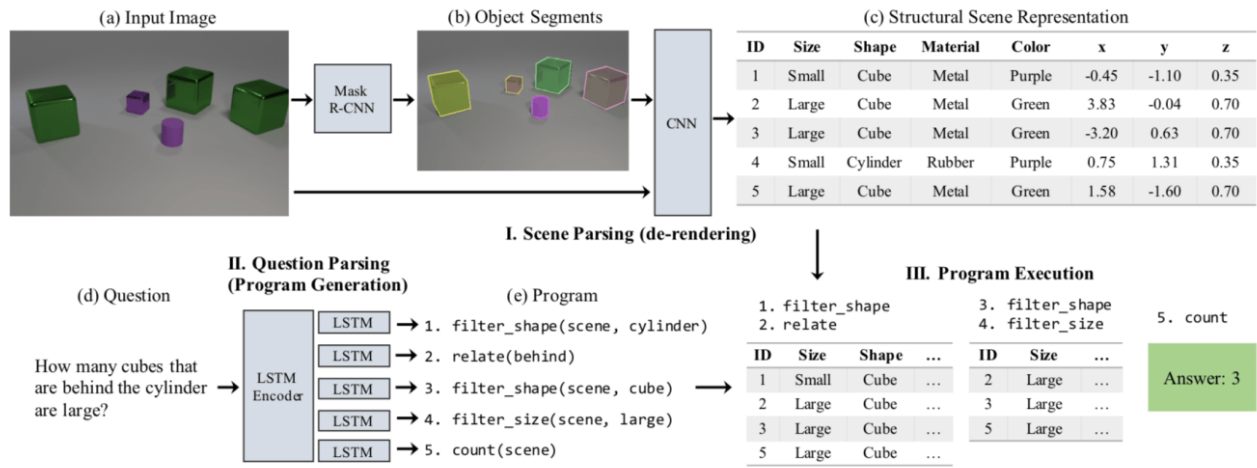

Figure 11, Image learning and path planning with the algorithm

## Conclusion

We demonstrated a variety of controls paradigms within a single software hardware framework, including neuro-symbolic learning and controls, using rules and frames based reasoning, classic motion control, and various neural approaches to perception and controls. Combining several state of the art controllers with computer animation tools, various solutions of grasping tasks with force feedback, promises new investigations in machine learning with embodiment, to pursue multi-modal, multidimensional trained models that more closely correlate with the human experience.

The results show clear promise for short term social utility in arts and collaborative robotic applications, by signaling with intuitive social communication of humanlike form and motion, both as interactive fiction and embodied social agent. We hope that continued experiments combining symbolic and neural AI, on a social robot platform and with robust simulation tools, may produce new opportunities for experiments in embodied cognition, and help progress towards more generalized intelligence in machines. We also believe such creative embodied neuro-cognitive research may open new lines of research in self learning and accelerate the pursuit of human-level AI.